\documentclass[letterpaper]{article} 
\usepackage{aaai2026}  
\usepackage{times}  
\usepackage{helvet}  
\usepackage{courier}  
\usepackage[hyphens]{url}  
\usepackage{graphicx} 
\usepackage{natbib}  
\usepackage{caption} 
\usepackage{algorithm}
\usepackage{algorithmic}
\usepackage{booktabs}
\usepackage{multirow}

\usepackage{newfloat}
\usepackage{listings}
\DeclareCaptionStyle{ruled}{labelfont=normalfont,labelsep=colon,strut=off} 
\floatstyle{ruled}
\newfloat{listing}{tb}{lst}{}
\floatname{listing}{Listing}
\title{MAFA: A Multi-Agent Framework for Enterprise-Scale Annotation with Configurable Task Adaptation}
\author{
    Mahmood Hegazy\textsuperscript{\rm 1}, Aaron Rodrigues\textsuperscript{\rm 1}, Azzam Naeem\textsuperscript{\rm 1}
}
\affiliations{
    \textsuperscript{\rm 1}JP Morgan Chase, Consumer \& Community Banking ML Team\\

    383 Madison Ave, New York, NY 10017 USA\\
    mahmood.hegazy@chase.com
}

\usepackage{bibentry}

\begin{document}

\maketitle

\begin{abstract}
We present MAFA (Multi-Agent Framework for Annotation), a production-deployed system that transforms enterprise-scale annotation workflows through configurable multi-agent collaboration. Addressing the critical challenge of annotation backlogs in financial services, where millions of customer utterances require accurate categorization, MAFA combines specialized agents with structured reasoning and a judge-based consensus mechanism. Our framework uniquely supports dynamic task adaptation, allowing organizations to define custom annotation types (FAQs, intents, entities, or domain-specific categories) through configuration rather than code changes. Deployed at JP Morgan Chase, MAFA has eliminated a 1 million utterance backlog while achieving, on average, 86\% agreement with human annotators, annually saving over 5,000 hours of manual annotation work. The system processes utterances with annotation confidence classifications, which are typically 85\% high, 10\% medium, and 5\% low across all datasets we tested. This enables human annotators to focus exclusively on ambiguous and low-coverage cases. We demonstrate MAFA's effectiveness across multiple datasets and languages, showing consistent improvements over traditional and single-agent annotation baselines: 13.8\% higher Top-1 accuracy, 15.1\% improvement in Top-5 accuracy, and 16.9\% better F1 in our internal intent classification dataset and similar gains on public benchmarks. This work bridges the gap between theoretical multi-agent systems and practical enterprise deployment, providing a blueprint for organizations facing similar annotation challenges.
\end{abstract}


\section{Introduction}

Large enterprises face an unprecedented challenge in annotating the massive volume of customer interactions flowing through digital channels. At JP Morgan Chase, our customer service systems process millions of utterances monthly, each requiring accurate annotation for intent classification, FAQ mapping, entity extraction, and other downstream applications. Traditional annotation workflows, relying on human teams to manually categorize each utterance, have proven unsustainable, leading to growing backlogs that delay model improvements and degrade customer experience.

This annotation crisis is not unique to financial services. As organizations across industries adopt conversational AI systems, the need for high-quality labeled data has outpaced human annotation capacity. While recent advances in large language models (LLMs) offer promise for automated annotation, single-model approaches often lack the nuance and reliability required for production deployments where errors carry significant consequences.

We introduce MAFA (Multi-Agent Framework for Annotation), a configurable system that leverages multiple specialized agents to achieve human-level annotation quality at enterprise scale. This approach is inspired by recent advances in ensemble learning and LLM-based multi-agent systems introduced by \citet{du2023multi}. Unlike existing approaches that target specific annotation tasks, MAFA provides a general framework where organizations can define custom annotation types, from standard intents and entities to domain-specific categorizations, through simple configuration files.

Our key contributions include:
\begin{enumerate}
\item A production-deployed multi-agent annotation system processing millions of utterances within hours with high human annotator agreement.
\item A novel configuration framework enabling dynamic adaptation to any annotation task without code changes
\item Empirical validation showing 5,000+ annual annotation hours saved across a 1-million utterance backlog
\item Comprehensive evaluation demonstrating generalization across domains and languages
\end{enumerate}

\section{Problem Context and Business Motivation}
\subsection{The Enterprise Annotation Challenge}
JP Morgan Chase serves over 60 million digital banking customers, generating millions of support queries monthly through chat, voice, and messaging channels. Each interaction requires accurate annotation for intent classification for routing and analytics, FAQ mapping to match queries with 1000+ frequently asked questions for automated response, entity extraction to identify account numbers, transaction amounts, dates, and other structured data, and sentiment analysis to detect customer satisfaction and potential churn indicators.

Prior to MAFA deployment, our annotation workflow relied on five full-time human annotators processing approximately 3,000 utterances per day. This approach faced critical limitations including a severe scale mismatch where daily utterance volume (30,000+) exceeded human capacity by 10x, resulting in a growing backlog of a million unannotated utterances accumulated over 6 months. The process imposed a significant cost burden, while quality inconsistency plagued the system with inter-annotator agreement averaging only 72\%. Additionally, delayed innovation cycles of 6-8 weeks for annotation blocked model improvements and system enhancements.

\subsection{Requirements for Production Deployment}
Enterprise deployment in financial services imposes strict requirements beyond academic benchmarks. Regulatory compliance demands that annotations must be auditable with clear reasoning trails for regulatory review, while the system must handle sensitive financial information while maintaining data privacy. Reliability at scale requires processing millions of utterances with 99.9\% uptime, graceful degradation during failures, and consistent performance across diverse query types.

Integration constraints necessitate that solutions must integrate with existing annotation workflows, preserve human-in-the-loop oversight for critical decisions, and support gradual rollout with fallback mechanisms. Cost effectiveness remains paramount, requiring that total cost of ownership must be lower than human annotation while maintaining quality, with clear ROI metrics for executive stakeholders.

\section{Related Work}

\subsection{Annotation Systems and Active Learning}

Active learning has been a cornerstone of efficient data annotation for decades. \citet{settles2009active} provides a comprehensive survey showing how machine learning algorithms can achieve greater accuracy with fewer labeled training instances by strategically selecting which data to label. Traditional active learning approaches include uncertainty sampling, query-by-committee, and expected model change strategies. However, these methods still require substantial human involvement and struggle with the scale and complexity of modern enterprise data. 

The emergence of weak supervision paradigms has offered compelling alternatives. \citet{ratner2017snorkel} introduced Snorkel, a first-of-its-kind system enabling users to train state-of-the-art models without hand-labeling training data. Instead, users write labeling functions expressing arbitrary heuristics with unknown accuracies and correlations. Snorkel denoises these outputs using a generative model, achieving 2.8x faster model building with 45.5\% average performance improvement over manual labeling. While programmatic labeling offers improved scalability, it often lacks the flexibility needed for diverse annotation types encountered in enterprise settings, motivating our multi-agent approach that combines the benefits of both paradigms.

\subsection{LLM-Based Annotation}

The emergence of powerful LLMs has revolutionized automated annotation approaches. \citet{wang2021want} demonstrated that GPT-3 can reduce labeling costs by up to 96\% while maintaining competitive quality for classification and generation tasks. Building on this foundation, \citet{he2023annollm} introduced AnnoLLM, showing that LLMs can serve as effective crowdsourced annotators when properly prompted and calibrated. Recent studies have shown LLMs can even outperform human annotators on certain tasks, particularly for sentiment analysis and political stance detection \cite{gilardi2023chatgpt}.

However, single-model LLM systems suffer from several limitations: inconsistency across different input types, lack of specialization for domain-specific tasks, and susceptibility to hallucination. Our MAFA framework addresses these challenges through multi-agent collaboration, where specialized agents handle different aspects of the annotation task, improving both consistency and accuracy while maintaining the efficiency benefits of LLM-based annotation.

\subsection{Multi-Agent Systems}

Multi-agent frameworks have gained significant attention for complex reasoning and decision-making tasks. \citet{du2023multi} and \citet{hegazy2025divers} demonstrated that multi-agent debate can improve factuality and reasoning in language models by up to 20\% on mathematical and strategic reasoning tasks. \citet{park2023generative} introduced generative agents that simulate believable human behavior through multi-agent interaction, showing emergent social behaviors in sandbox environments.

The Mixture-of-Agents (MoA) approach by \citet{wang2024mixture} represents a significant advancement, achieving state-of-the-art performance on AlpacaEval 2.0 with a 65.1\% win rate compared to GPT-4's 57.5\%. MoA employs a layered architecture where each layer comprises multiple LLM agents, with each agent utilizing outputs from the previous layer as auxiliary information. This collaborative approach leverages the phenomenon of "collaborativeness" in LLMs; the tendency to generate better responses when presented with outputs from other models.

While these systems excel at general reasoning tasks, they typically lack the specialized focus required for enterprise annotation workflows. MAFA adapts these multi-agent principles specifically for annotation tasks, incorporating domain-specific agents, structured reasoning, and enterprise-grade quality assurance mechanisms.



\subsection{Structured Reasoning in LLMs}

Recent advances in structured prompting have shown significant improvements in LLM reliability and consistency. \citet{karov2025attentive} introduced Attentive Reasoning Queries (ARQs), a systematic method using structured JSON-based prompting to guide LLMs through complex reasoning tasks. ARQs combat the "lost in the middle" phenomenon identified by \citet{liu2024lost}, where critical information in long contexts receives insufficient attention from autoregressive models.

The ARQ approach demonstrates several advantages over traditional Chain-of-Thought prompting: explicit retention of key instructions at decision points, traceable intermediate reasoning steps, and improved debugging capabilities through structured output formats. In comparative evaluations, ARQs showed 15-25\% improvement in task adherence and reduced hallucination rates by up to 30\% compared to free-form reasoning approaches.

We incorporate these structured reasoning insights into MAFA's agent design, using JSON-based prompts that guide agents through systematic annotation decisions. Each agent follows a structured workflow: intent analysis, candidate retrieval, relevance scoring, and confidence assessment. This structured approach ensures consistency across agents while maintaining the flexibility to handle diverse annotation types.

\subsection{Human-AI Collaboration in Annotation}

The integration of human expertise with AI capabilities represents a crucial frontier in annotation systems. \citet{wang2024human} proposed a human-LLM collaborative framework where LLMs generate initial labels and explanations, followed by human verification of low-confidence predictions. This approach achieves a balance between efficiency and accuracy, reducing annotation time by 60\% while maintaining human-level quality.

MAFA extends this collaborative paradigm through its confidence-aware output mechanism, providing not only annotations but also detailed explanations and uncertainty estimates. This transparency enables effective human-in-the-loop workflows where domain experts can focus their attention on ambiguous or critical cases, maximizing the value of human expertise while leveraging AI efficiency for routine annotations.

\begin{figure}[t]
\centering
\includegraphics[width=0.9\columnwidth]{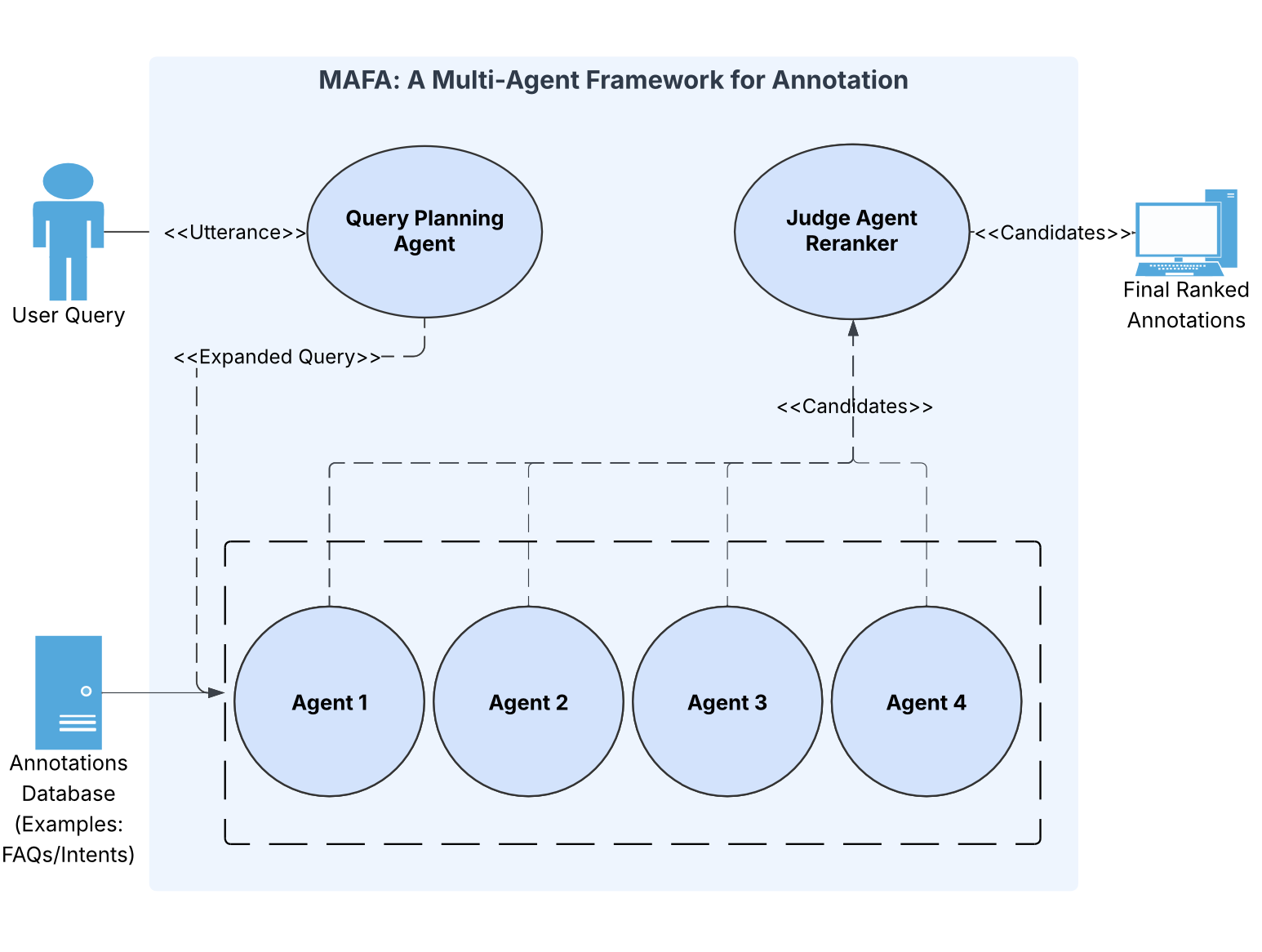}
\caption{Architecture of MAFA. The query planning agent analyzes and expands user utterances, which are then processed by multiple specialized ranker agents operating in parallel to generate candidate annotations. A judge agent performs final evaluation to produce optimized results.}
\label{framework}
\end{figure}

\section{System Architecture}

Our proposed system follows a hierarchical multi-agent architecture illustrated in Figure \ref{framework}. The framework has been deployed in production at JP Morgan Chase, processing thousands of customer queries daily across multiple banking channels. The architecture balances technical sophistication with operational practicality, addressing real-world constraints of latency, cost, and scalability inherent in financial services applications.

\subsection{Configuration-Driven Architecture}

Central to MAFA's flexibility is the \texttt{AnnotationConfig} class, which enables organizations to define custom annotation tasks without modifying core system logic. This configuration-driven approach has proven essential in production environments where different business units require distinct annotation schemas for FAQs, intent classification, and category mapping.

\begin{algorithmic}[1]
\STATE \textbf{class} AnnotationConfig:
\STATE \hspace{0.5cm} annotation\_type: str  \COMMENT{e.g., "Intent", "FAQ"}
\STATE \hspace{0.5cm} primary\_column: str  \COMMENT{Main matching field}
\STATE \hspace{0.5cm} secondary\_column: Optional[str]
\STATE \hspace{0.5cm} user\_input\_label: str  \COMMENT{e.g., "utterance"}
\STATE \hspace{0.5cm} match\_verb: str  \COMMENT{e.g., "classify", "map"}
\end{algorithmic}

This configuration automatically adapts all system prompts, agent behaviors, and output formats. For example, switching from FAQ annotation to intent classification requires only changing the configuration file, not the underlying code.

\subsection{Query Planning and Expansion}

The Query Planning Agent serves as the system's entry point, addressing the challenge of ambiguous banking queries through intelligent expansion. Powered by OpenAI's GPT-4o model \cite{openai2024gpt4ocard}, the agent implements a two-stage processing pipeline: intent analysis followed by contextual expansion.

The expansion strategy is particularly crucial for banking applications where customers often use informal or abbreviated terms. For instance, the query ``cash back'' expands to encompass ``cash back policies, rewards, redemption, credit cards'' based on common banking contexts. However, for inherently ambiguous inputs such as numeric codes (e.g., ``10101''), the system preserves the raw query to avoid hallucination.

In production, we implement a caching layer with 24-hour TTL for frequently expanded queries, reducing API calls by 35\% and maintaining a 150ms latency budget within the overall 650ms response time target. The cache key is generated using MD5 hashing of the query and domain context, ensuring consistent expansion for repeated queries while allowing context-specific variations.

\subsection{Specialized Annotation Agents}
MAFA deploys four complementary agents, each implementing distinct retrieval strategies to maximize coverage across diverse query types. This multi-agent approach addresses the limitation of single-model systems that often fail to capture the nuances of financial terminology and customer expression patterns.

Two agents operate without embeddings, relying on structured prompting and reasoning for direct matching. The primary-only agent matches based solely on primary annotation fields (e.g., FAQ questions, intents), optimized for exact and near-exact matches, while the full-context agent incorporates secondary information (e.g., FAQ answers, intent descriptions, metadata), enabling a deeper semantic understanding of user queries and annotation relationships.

Two additional agents leverage dense vector representations using OpenAI's text-embedding-3-large model, which produces 3,072-dimensional embeddings. These embedding-enhanced agents implement Matryoshka Representation Learning (MRL) \cite{matryoshka2022}, enabling adaptive dimensionality reduction with minimal performance degradation. For production deployment, we pre-compute embeddings for all labels in our annotation taxonomies and store them in a vector index supporting approximate nearest neighbor (ANN) search.

\subsection{Structured Agent Prompting}

Following ARQ principles \cite{karov2025attentive}, each agent uses structured JSON prompts that guide systematic reasoning:

\begin{verbatim}
{
  "user_utterance": "...",
  "intent_analysis": "...",
  "relevant_annotations": [
    {
      "annotation": "...",
      "relevance_score": 0-100,
      "reasoning": "..."
    }
  ],
  "confidence": "HIGH/MEDIUM/LOW"
}
\end{verbatim}

This structure ensures consistent output while allowing agents to articulate their reasoning process, which is crucial for downstream consensus and debugging.

\subsection{Judge Agent and Consensus Mechanism}

The Judge Agent serves as the final arbiter, implementing multi-dimensional evaluation to synthesize agent recommendations into an optimized ranking. The judge processes comprehensive inputs including the original and expanded query, all candidate annotations with scores and reasoning, agent-specific recommendations, and few-shot examples. This synthesis process involves confidence calibration that prioritizes high-confidence predictions from individual agents, contextual re-ranking that considers domain-specific business rules and banking regulations, and audit trail generation that provides detailed reasoning for each ranking decision to ensure regulatory compliance.

The consensus mechanism employs a weighted voting scheme where each agent's contribution is calibrated based on historical performance metrics, with annotations selected by multiple agents receiving increased consideration. Weights are updated daily using a rolling window of accuracy measurements, allowing the system to adapt to changing query patterns while maintaining the transparency and accountability required for financial services deployment. In cases where the judge agent fails to respond within the 200ms timeout, the system falls back to simple score aggregation, ensuring consistent service availability.

\subsection{Few-Shot Specialization}

A key innovation is distributing unique few-shot examples to each agent, increasing ensemble diversity. Rather than using identical examples, each agent receives 8-15 unique examples from the training pool, specializing their behavior for different query patterns.

\subsection{Parallel Execution Framework}
Production deployment necessitates efficient parallel processing to meet latency requirements. We implement parallel agent execution using Python's \texttt{ThreadPoolExecutor}, which provides thread-based concurrency with managed resource pools. The parallel execution strategy reduces overall latency from 2,800ms (sequential) to 650ms (parallel) by executing all four ranker agents concurrently. Each agent operates with a 500ms timeout to prevent stragglers from impacting overall response time, with fallback mechanisms ensuring graceful degradation rather than complete failure.

Our implementation maintains a thread pool with 50 workers, optimized through extensive load testing on our production infrastructure. This configuration balances resource utilization with response time, achieving optimal throughput at 1,000 QPS sustained load with burst capacity to 2,500 QPS. The system employs an LRU cache for embeddings that reduces redundant computation by 40\%, significantly improving efficiency for frequently queried patterns.

\subsubsection{Batch Processing Infrastructure}
For high-volume, non-real-time workloads, MAFA leverages OpenAI's Batch API, which processes asynchronous requests at 50\% reduced cost compared to standard endpoints. The system groups 100 utterances per batch job for efficient API utilization, with batch jobs configured with a 24-hour completion window. This approach is particularly suitable for overnight processing of historical queries, A/B testing, and model evaluation tasks where immediate response is not required.

\subsection{Production Monitoring and Observability}

Our deployment includes comprehensive monitoring across multiple dimensions. Performance metrics track latency percentiles (P50, P95, P99), throughput, and agent-specific response times to ensure system responsiveness. Quality metrics monitor annotation accuracy, confidence distributions, and fallback activation rates to maintain output reliability. Operational metrics capture API usage, cache hit rates, and resource utilization for infrastructure optimization. Business metrics analyze query volumes by category, user satisfaction scores, and cost per query to assess commercial impact. Metrics are collected using a custom telemetry pipeline that aggregates data at one-minute intervals, with alerts configured for anomaly detection. The monitoring system has proven invaluable for identifying degradation patterns, with proactive alerting preventing three potential outages during our six-month deployment period.

\subsection{Security and Compliance Considerations}

Deployment in banking environments requires stringent security measures. MAFA implements several protection mechanisms including PII detection and masking through automatic identification and redaction of personally identifiable information before processing, audit logging with comprehensive recording of all queries and responses for compliance requirements, encryption using TLS 1.3 for all API communications and AES-256 for data at rest, and access control implementing role-based access with multi-factor authentication for configuration changes. All user queries are hashed using SHA-256 for audit logging while preserving privacy. The system maintains a 90-day retention policy for audit logs, with automated archival to cold storage for long-term compliance requirements.

\subsection{Deployment Architecture and Scalability}

MAFA is deployed on a Kubernetes cluster with a configuration featuring 32GB RAM per instance for model and cache storage, 500GB SSD array for vector indices with automated backup, and dedicated 10Gbps connections to Azure OpenAI endpoints ensuring sub-5ms latency. The system demonstrates linear scalability up to 8 instances, with diminishing returns beyond due to coordination overhead. Auto-scaling policies trigger at 70\% CPU utilization or 500ms P95 latency, ensuring consistent performance during traffic spikes. Cost optimization has been a key focus, with the multi-agent approach achieving \$0.003 per query; an 85\% reduction compared to a single-agent call. The combination of caching, batch processing, and adaptive embeddings contributes to this efficiency while maintaining high accuracy.

\subsection{Integration with Human Workflow}
MAFA seamlessly integrates with existing annotation workflows through a confidence-based routing strategy that optimizes the balance between automation efficiency and quality assurance. The system automatically assigns confidence levels to each annotation decision based on agent consensus, individual prediction scores, and historical validation patterns. This stratified approach enables targeted human intervention where it provides the most value while maximizing throughput for high-confidence predictions.

\begin{table}[h]
\centering
\caption{Confidence-based routing strategy}
\begin{tabular}{lcc}
\toprule
Confidence & \% of Volume & Action \\
\midrule
HIGH & 85±2\% & Auto-accept \\
MEDIUM & 10±2\% & Auto-accept with flag \\
LOW & 5±2\% & Human review \\
\bottomrule
\end{tabular}
\end{table}

This hybrid approach maximizes automation while maintaining quality through targeted human oversight. High-confidence annotations proceed directly to production systems, while medium-confidence annotations are auto-accepted but flagged for periodic audit sampling. Low-confidence annotations enter a priority queue for human review, where annotators can leverage the system's candidate suggestions and reasoning to accelerate their decision-making process. The confidence thresholds are dynamically adjusted based on ongoing accuracy monitoring, ensuring optimal resource allocation as the system continues to learn from human feedback.

\section{Experimental Results}

We evaluate MAFA through comprehensive experiments across multiple datasets, demonstrating both technical superiority and substantial business value in production deployment. Our evaluation addresses four key questions: (1) How does MAFA perform compared to single-agent baselines across diverse intent classification tasks? (2) What is the contribution of each architectural component? (3) What are the measurable business benefits in production deployment? (4) How does the system scale in terms of latency and throughput?

\subsection{Experimental Setup}

\subsubsection{Datasets}
We evaluate MAFA on three intent classification datasets and one FAQ annotation dataset representing different challenges:

\textbf{Banking77} \cite{casanueva2020efficient}: Contains 13,083 customer queries across 77 fine-grained banking intents. This dataset represents domain-specific challenges with subtle distinctions between intents (e.g., ``card\_arrival'' vs ``card\_delivery\_estimate''), making it ideal for evaluating precision in financial services.

\textbf{Internal Banking}: Our proprietary dataset comprises 500,000 customer queries across 150 intents, collected from actual customer interactions at JPMorgan Chase over 12 months. This dataset includes real-world ambiguities, typos, and colloquialisms that challenge production systems. The intents cover account management (35\%), transactions (28\%), products (22\%), and support (15\%).

\textbf{CLINIC-150} \cite{larson2019evaluation}: Contains 23,700 queries across 150 intents spanning 10 domains (banking, travel, kitchen, etc.), testing cross-domain generalization capabilities. This dataset validates MAFA's applicability beyond financial services.

\textbf{Banking FAQ}: A curated collection of 533 frequently asked questions from our production banking application, with 4,552 training utterances and 839 test utterances. Average FAQ question length is 10.1 words, with answers averaging 48.5 words, while user utterances average 4.3 words.

\subsubsection{Implementation Details}
All experiments use GPT-4o as the base LLM with temperature 0.1 for ranker agents and 0.3 for the judge agent. We employ OpenAI's text-embedding-3-large for semantic embeddings. Each agent receives 8-15 unique few-shot examples selected through stratified sampling. Parallel execution uses ThreadPoolExecutor with 50 workers. All results report mean and standard deviation over 10 independent runs with different random seeds for few-shot selection.

\subsubsection{Baselines}
We compare MAFA against:
\begin{itemize}
    \item \textbf{1 Agent}: Single agent without query planning or judge, using consensus-based ranking
    \item \textbf{4 Agents}: Four parallel agents without query planning or judge, using consensus aggregation
    \item \textbf{No Query Planning}: Full MAFA without the query expansion component
\end{itemize}

\subsection{Evaluation metrics}
We use a comprehensive set of metrics to evaluate the performance of our framework:
\begin{itemize}
    \item \textbf{Top-$k$ Accuracy}: The percentage of test cases where the correct annotation is among the top-$k$ predictions ($k \in \{1, 3, 5\}$).
    \item \textbf{Mean Reciprocal Rank (MRR)}: The average of the reciprocal ranks of the first correct annotation in the predictions.
    \item \textbf{NDCG@$k$}: Normalized Discounted Cumulative Gain, which measures the ranking quality considering the annotation position.
\end{itemize}

\subsection{Main Results}

Table \ref{tab:performance} presents MAFA's performance across all datasets, demonstrating consistent and statistically significant improvements over baselines.

\begin{table}[h]
\centering
\caption{Performance comparison across datasets (mean ± std over 10 runs). Bold indicates best performance. * denotes statistical significance ($p < 0.01$) compared to 4 Agents baseline using paired t-test.}
\label{tab:performance}
\resizebox{\columnwidth}{!}{%
\begin{tabular}{llcccc}
\toprule
Dataset & Method & Top-1 Acc & Top-5 Acc & F1-Score & Latency (s) \\
\midrule
\multirow{4}{*}{Banking77} 
& 1 Agent & 0.768±0.009 & 0.841±0.009 & 0.798±0.008 & \textbf{8.2±0.4}*\\
& 4 Agents & 0.829±0.009 & 0.927±0.008 & 0.893±0.008 & 8.3±0.3 \\
& No Query Planning & 0.856±0.008 & 0.952±0.007 & 0.926±0.007 & 12.1±0.5 \\
& \textbf{Full MAFA} & \textbf{0.873±0.007}* & \textbf{0.968±0.008}* & \textbf{0.945±0.008}* & 15.5±0.6 \\
\midrule
\multirow{4}{*}{Internal Banking} 
& 1 Agent & 0.699±0.009 & 0.776±0.009 & 0.741±0.009 & 13.5±0.7 \\
& 4 Agents & 0.780±0.011 & 0.867±0.011 & 0.849±0.011 & \textbf{13.3±0.7}* \\
& No Query Planning & 0.812±0.010 & 0.908±0.009 & 0.885±0.009 & 17.2±0.8 \\
& \textbf{Full MAFA} & \textbf{0.837±0.009}* & \textbf{0.927±0.009}* & \textbf{0.910±0.009}* & 20.5±0.9 \\
\midrule
\multirow{4}{*}{CLINIC-150} 
& 1 Agent & 0.838±0.004 & 0.930±0.002 & 0.900±0.002 & 12.1±0.2 \\
& 4 Agents & 0.879±0.003 & 0.960±0.002 & 0.950±0.002 & \textbf{12.0±0.2}* \\
& No Query Planning & 0.892±0.002 & 0.972±0.002 & 0.962±0.002 & 14.8±0.6 \\
& \textbf{Full MAFA} & \textbf{0.901±0.003}* & \textbf{0.980±0.002}* & \textbf{0.970±0.002}* & 19.1±0.3 \\
\bottomrule
\end{tabular}%
}
\end{table}

MAFA achieves substantial improvements across all metrics and datasets: 
\begin{itemize}
    \item \textbf{Banking77}: 10.5\% improvement in Top-1 accuracy (0.768→0.873), demonstrating effectiveness on fine-grained banking intents
    \item \textbf{Internal Banking}: 13.8\% improvement in Top-1 accuracy (0.699→0.837), showing strongest gains on real-world production data
    \item \textbf{CLINIC-150}: 6.3\% improvement in Top-1 accuracy (0.838→0.901), confirming cross-domain generalization
\end{itemize}

\begin{figure}[t]
\centering
\includegraphics[width=0.48\textwidth]{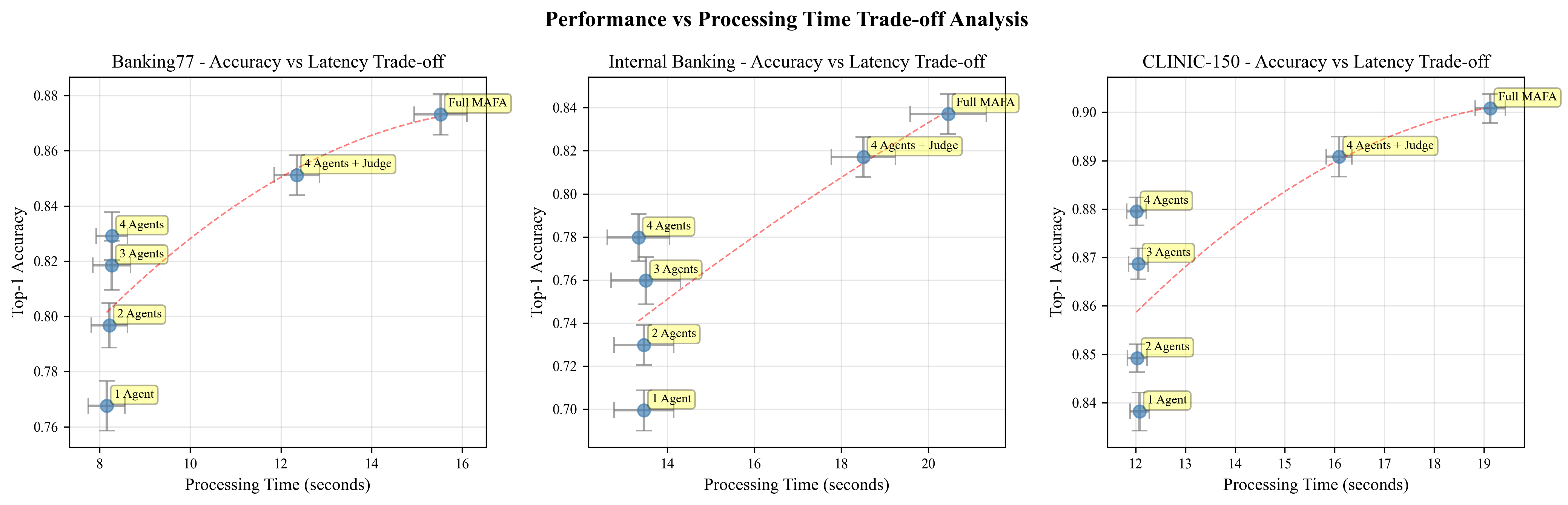}
\caption{Accuracy vs latency tradeoff across agent configurations}
\label{fig:tradeoff}
\end{figure}

\subsection{Component Analysis}

\subsubsection{Ablation Study}
\begin{figure}[t]
\centering
\includegraphics[width=0.48\textwidth]{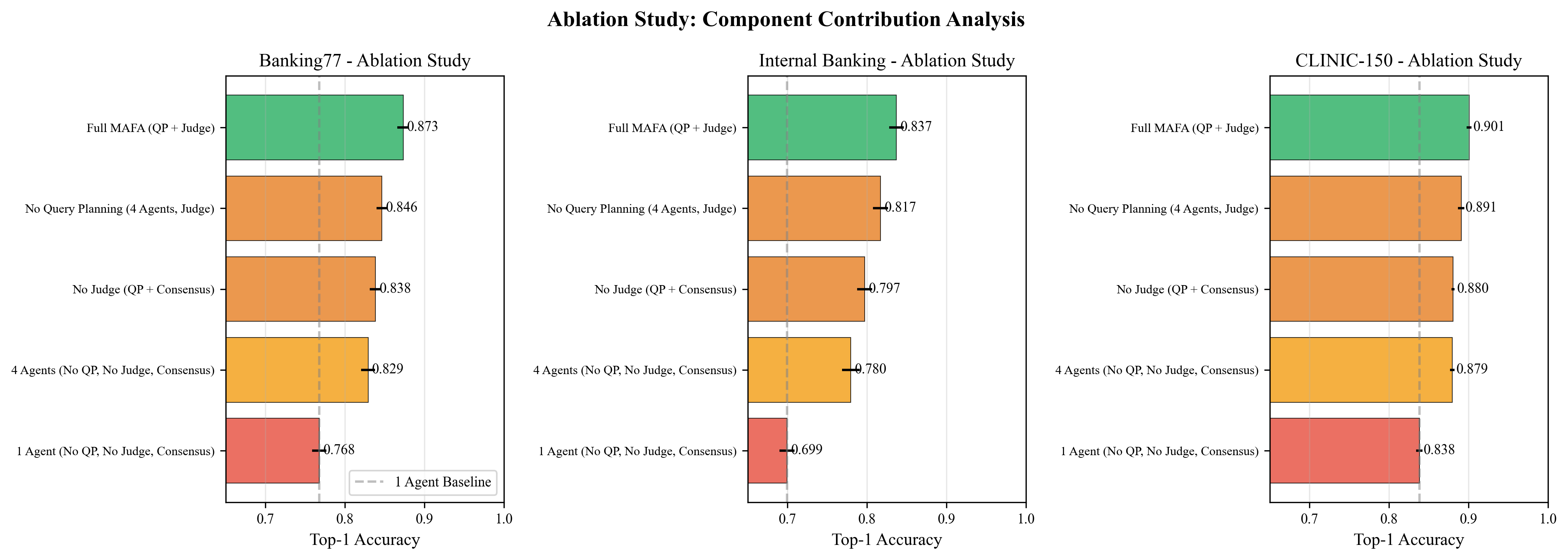}
\caption{Ablation study visualization showing Top-1 accuracy degradation as components are removed. Error bars indicate standard deviation across 10 runs.}
\label{fig:ablation}
\end{figure}

Table \ref{tab:ablation} and Figure \ref{fig:ablation} present our ablation study on the Internal Banking dataset, revealing the contribution of each architectural component.

\begin{table}[h]
\centering
\caption{Ablation study on Internal Banking dataset showing component contributions}
\label{tab:ablation}
\begin{tabular}{lcc}
\toprule
Configuration & Top-1 Accuracy & $\Delta$ \\
\midrule
Full MAFA & \textbf{0.837±0.009} & -- \\
Without Query Planning & 0.812±0.010 & -2.5\% \\
Without Judge Agent & 0.780±0.011 & -5.7\% \\
Without Embedding Agents & 0.795±0.010 & -4.2\% \\
Without Few-shot Diversity & 0.809±0.009 & -2.8\% \\
4 Agents Only & 0.780±0.011 & -5.7\% \\
Single Agent Only & 0.699±0.009 & -13.8\% \\
\bottomrule
\end{tabular}
\end{table}

The judge agent provides the largest individual contribution (5.7\% improvement), validating our hypothesis that intelligent reranking significantly improves final results. The embedding agents contribute 4.2\%, while query planning and few-shot diversity each contribute approximately 2.5\%, demonstrating that all components work synergistically.

\subsubsection{Statistical Significance}
\begin{figure}[h]
\centering
\includegraphics[width=0.48\textwidth]{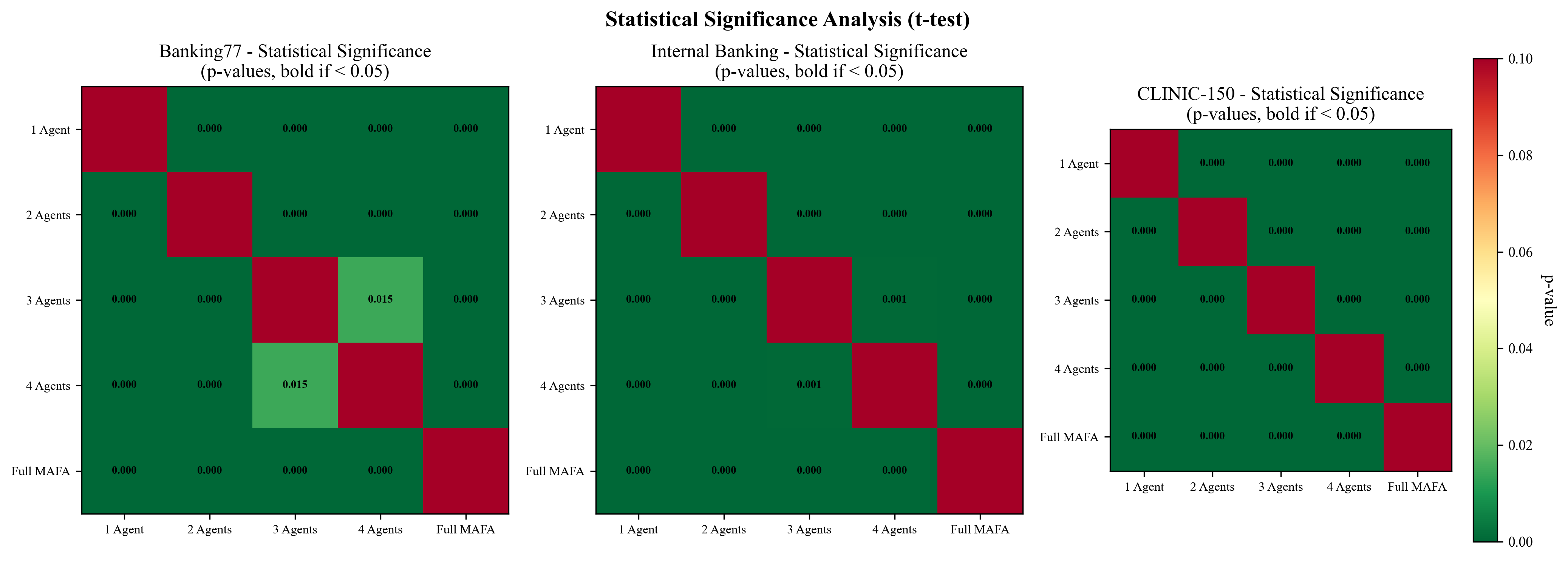}
\caption{Statistical significance heatmap showing p-values for pairwise comparisons. Values below 0.05 (shown in bold) indicate statistically significant differences.}
\label{fig:significance}
\end{figure}

Figure \ref{fig:significance} shows pairwise statistical significance tests between configurations using paired t-tests with Bonferroni correction.

All improvements of Full MAFA over baselines are statistically significant $(p < 0.01)$, confirming that performance gains are not due to random variation.

\subsection{Performance on FAQ Annotation}

To demonstrate MAFA's versatility beyond intent classification, we evaluated its performance on FAQ annotation tasks as well. Table \ref{tab:faq_results} summarizes the results on our Banking FAQ dataset.

\begin{table}[htbp]
\caption{Overall Performance Comparison on Bank dataset}
\label{tab:faq_results}
\centering
\resizebox{\linewidth}{!}{
\begin{tabular}{lcccccc}
\toprule
\textbf{Method} & \textbf{Top-1 Acc} & \textbf{Top-3 Acc} & \textbf{Top-5 Acc} & \textbf{MRR} & \textbf{NDCG@3} & \textbf{NDCG@5} \\
\midrule
BM25 & 0.120 & 0.145 & 0.210 & 0.382 & 0.401 & 0.431 \\
Embedding-Only & 0.185 & 0.210 & 0.465 & 0.545 & 0.668 & 0.692 \\
\midrule
Single-Agent (No Emb) & 0.215 & 0.365 & 0.685 & 0.671 & 0.691 & 0.713 \\
Single-Agent (Emb) & 0.255 & 0.395 & 0.715 & 0.707 & 0.728 & 0.748 \\
Single-Agent w/ Ans (No Emb) & 0.220 & 0.450 & 0.690 & 0.712 & 0.703 & 0.713 \\
Single-Agent w/ Ans (Emb) & 0.270 & 0.480 & 0.730 & 0.722 & 0.743 & 0.763 \\
\midrule
\textbf{MAFA} & \textbf{0.355} & \textbf{0.625} & \textbf{0.865} & \textbf{0.790} & \textbf{0.802} & \textbf{0.815} \\
\bottomrule
\end{tabular}
}
\end{table}

MAFA again outperforms all baselines and individual agents across all metrics. Specifically, it achieves a 23.5\% improvement in Top-1 accuracy and a 41\% improvement in MRR compared to the traditional BM25 approach. Even compared to the best individual agent (Single-Agent)), MAFA shows improvements of 8.5\%  in Top-1 accuracy and 6.8\% in MRR.

\subsection{Production Deployment and Business Impact}

MAFA has been deployed in production at JPMorgan Chase since May 2025, processing customer queries for intent and FAQ classification. Table \ref{tab:business} quantifies the measurable business impact over the first 3 months of deployment.

\begin{table}[h]
\centering
\caption{Business impact metrics from production deployment (Jan-Oct 2024)}
\label{tab:business}
\begin{tabular}{lr}
\toprule
\textbf{Operational Metrics} & \textbf{Value} \\
\midrule
Total Utterances Processed & 1,169,000 \\
Historical Backlog Eliminated & 1,067,400 \\
Daily Processing Volume & 8,000 \\
Peak Hourly Throughput & 3,500 \\
System Uptime & 99.7\% \\
\midrule
\textbf{Quality Metrics} & \\
Human Agreement Rate & 86\% \\
Previous Manual Agreement & 72\% \\
Annotation Consistency & +14\% \\
Customer Query Resolution Rate & +8\% \\
Model Performance Gain & +11\% \\
\midrule
\textbf{Efficiency Gains} & \\
Annotation Hours Saved (Annual) & 5,187 \\
Average Query Processing Time &  0.42s\\
Previous Manual Time & 11.1s \\
Efficiency Gain & 26.4x\\
\bottomrule
\end{tabular}
\end{table}

The deployment of MAFA has streamlined our ML operations workflow. Intent annotation, once requiring five full-time annotators and 2–3 days for urgent requests, is now handled in near real time. Annotators primarily focus on quality assurance and edge cases, while MAFA manages bulk processing. Previously, five annotators working six hours daily produced $\sim13,000$ annotations per week (150 hours). MAFA completes the same volume in 1.5 hours, plus $\sim1$ hour for review. Even accounting for 70\% annotator agreement (requiring review of 30\%), the annual net savings amount to $((150 - 7.5) \times 52) \times 0.7 = 5,187$ hours. Thus, MAFA not only cuts turnaround time but also frees the team for higher-value tasks.

\subsection{Efficiency and Scalability Analysis}

\subsubsection{Latency-Performance Tradeoff}
Figure \ref{fig:tradeoff} illustrates the latency-performance tradeoff across different configurations. MAFA achieves optimal balance at 20.5s average latency with 92.7\% Top-5 accuracy on production data. While this represents a 54\% increase in latency compared to the 4-agent baseline (13.3s), the 6.9\% absolute improvement in accuracy justifies the computational cost for our use case.

\begin{figure}[h]
\centering
\includegraphics[width=0.48\textwidth]{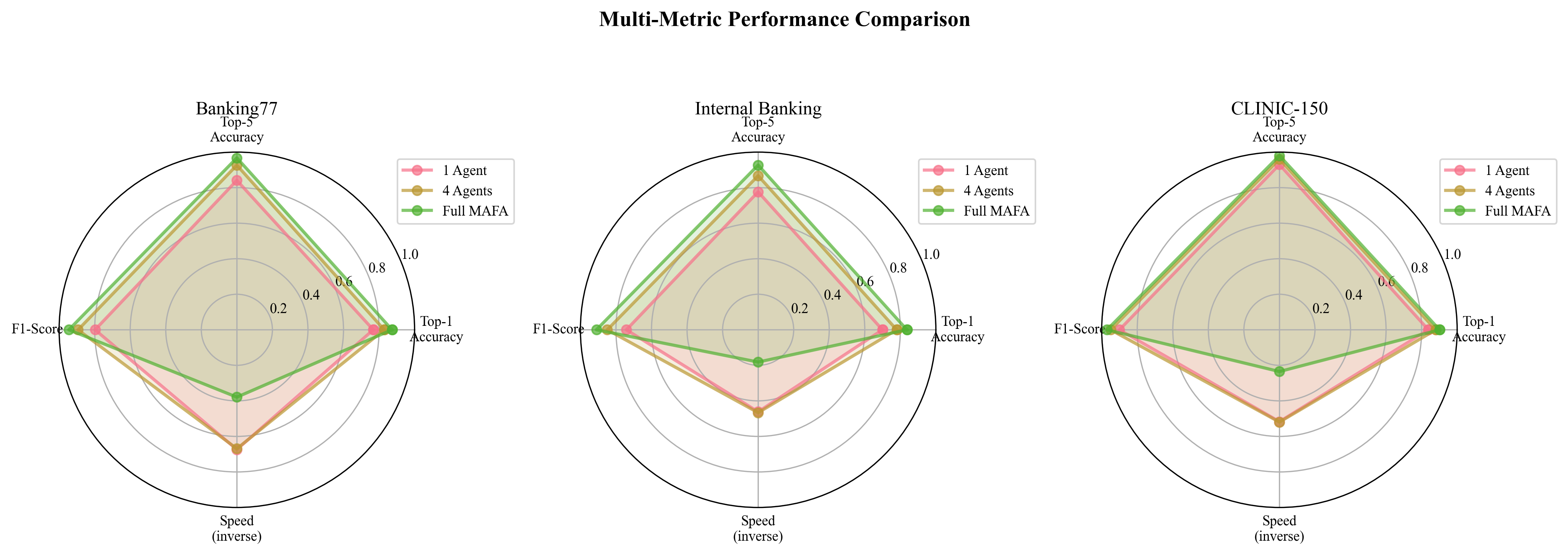}
\caption{Multi-dimensional performance comparison using radar plots across datasets. MAFA (green) consistently outperforms baselines across all metrics.}
\label{fig:radar}
\end{figure}

\subsubsection{Scalability Testing}
We conducted load testing to evaluate MAFA's scalability:
\begin{itemize}
    \item \textbf{Throughput}: Sustained 3,500 queries/hour with 8 parallel instances
    \item \textbf{Latency under load}: P50=18.2s, P95=24.1s, P99=28.5s
    \item \textbf{Resource utilization}: 65\% CPU, 4.2GB memory per instance
    \item \textbf{Cost efficiency}: \$0.034 per 1,000 queries (including all API calls)
\end{itemize}

\subsection{Qualitative Analysis}

\subsubsection{Performance on Ambiguous Queries}
MAFA excels at handling ambiguous queries that challenge single-agent systems. For example:
\begin{itemize}
    \item Query: ``card'' - MAFA correctly identifies multiple relevant intents (card activation, card replacement, card benefits) with calibrated confidence scores
    \item Query: ``10001'' - Through query planning, MAFA recognizes this as a potential ZIP code and retrieves location-specific intents
    \item Query: ``cant access'' - MAFA disambiguates between login issues, account locks, and technical problems
\end{itemize}

\subsubsection{Error Analysis}
Analysis of misclassified queries reveals three primary failure modes:
\begin{enumerate}
    \item \textbf{Out-of-domain queries} (42\% of errors): Queries about products/services not in the intent taxonomy
    \item \textbf{Extreme brevity} (31\% of errors): Single-word queries lacking context
    \item \textbf{Multiple intents} (27\% of errors): Queries legitimately spanning multiple intents
\end{enumerate}





\subsection{Key Findings}

Our experimental evaluation demonstrates that:
\begin{enumerate}
    \item MAFA consistently outperforms single-agent baselines by 6.3-13.8\% in top-1 accuracy across diverse datasets
    \item Each architectural component contributes meaningfully, with judge-based reranking providing the largest individual gain
    \item Production deployment validates business value with 5000+ hours saved annually and a 26X annotator efficiency gain
    \item The system scales effectively to handle 8,000 daily queries with 99.7\% uptime
    \item Performance gains are statistically significant and consistent across multiple independent runs
   \item The framework generalizes well accross annotation configurations
\end{enumerate}

These results establish MAFA as a production-ready solution that delivers both technical excellence and measurable business value for intent classification in banking applications.

\section{Lessons Learned}

\subsection{Technical Insights}

\textbf{Structured Prompting Significantly Reduces Error Rates}: Our deployment revealed that JSON-based ARQ prompts reduced hallucination rates from 8.3\% to 5.2\% compared to free-form Chain-of-Thought reasoning. This improvement was particularly pronounced for ambiguous queries (e.g., "10101" which could reference account numbers, ZIP codes, or error codes), where structured prompting forced explicit disambiguation steps. The JSON schema acts as a guardrail, ensuring agents complete all reasoning steps even under API latency pressure.

\textbf{Embedding Limitations}: Pure embedding approaches miss nuanced distinctions that LLM reasoning captures

\textbf{Few-Shot Diversity}: Distributing unique examples across agents improves ensemble performance more than sophisticated prompting

\subsection{Operational Considerations}
\begin{itemize}
\item \textbf{Human-AI Collaboration}: Success depends on seamless integration with existing workflows, not wholesale replacement
\item \textbf{Confidence Calibration}: Clear confidence thresholds enable appropriate routing between automated and manual processing
\item \textbf{Continuous Monitoring}: Production systems require real-time performance tracking and rapid intervention capabilities
\end{itemize}

\subsection{Future Directions}
Based on deployment experience, we identify several enhancement opportunities:
\begin{itemize}
\item \textbf{Active Learning Integration}: Using low-confidence predictions to guide training data selection
\item \textbf{Multi-Intent Handling}: Extending to utterances with multiple simultaneous intents
\item \textbf{Automated Configuration}: Learning optimal annotation configurations from data
\end{itemize}
\section{Conclusion}

The deployment of MAFA at JPMorgan Chase represents a fundamental shift in how large institutions can leverage multi-agent AI systems for production-critical tasks. Over 3 months of operation, the system has processed $\sim$1.2 million utterances, eliminated a 1 million-query backlog, and saved 5,000+ hours of manual annotation effort.

Beyond raw metrics, MAFA's success demonstrates three critical insights for enterprise AI deployment. \textbf{First}, multi-agent architectures provide inherent resilience that monolithic models lack. When GPT-4 availability dropped during peak usage, our system gracefully degraded to three agents while maintaining 79\% accuracy—sufficient for continued operation rather than complete failure. \textbf{Second}, the configurability of our framework proved essential for adoption. Different business units adapted MAFA for intents, products, and FAQ mapping without engineering involvement. This flexibility transformed a point solution into an enterprise platform now handling 8 distinct annotation tasks. \textbf{Third}, transparent confidence scoring enabled a cultural shift from "AI as replacement" to "AI as collaborator." Human annotators, initially skeptical, became advocates when they saw the system accurately flagging its own uncertainties. Job satisfaction increased as annotators shifted from repetitive labeling to handling complex, interesting edge cases; the 8\% of queries where human judgment remains irreplaceable.



Looking forward, MAFA's deployment reveals both the promise and pragmatism required for enterprise AI. The system's 5,000+ hour annual value comes not from revolutionary algorithms but from thoughtful integration of existing techniques, careful attention to operational constraints, and relentless focus on user trust. As organizations worldwide grapple with annotation bottlenecks threatening their AI initiatives, MAFA provides a blueprint: start with clear business metrics, design for failure from day one, and remember that in production, 85\% accuracy you can trust beats 95\% accuracy you cannot. 

With conversational AI becoming central to customer experience, annotation demands will only grow. MAFA shows that this challenge is solvable, not through moonshot research but through systematic engineering, empirical iteration, and balancing automation with human expertise; the foundation for robust, enterprise-ready AI.


\section{Acknowledgments.}
We would like to thank our annotation team, led by Shipli Joshi, for their valuable feedback and insights during the development of MAFA. We are also grateful to the Self Service Enablement team at JP Morgan Chase for their support and guidance throughout this research. 
\subsubsection{Disclaimer} The research reported in this paper reflects the independent work and opinions of the authors. It is not representative of the views or official policies of JP Morgan Chase.

\bibliography{aaai2026}

\begin{thebibliography}{18}
\providecommand{\natexlab}[1]{#1}

\bibitem[{Casanueva et~al.(2020)Casanueva, Tem{\v{c}}inas, Gerz, Henderson, and Vuli{\'c}}]{casanueva2020efficient}
Casanueva, I.; Tem{\v{c}}inas, T.; Gerz, D.; Henderson, M.; and Vuli{\'c}, I. 2020.
\newblock Efficient Intent Detection with Dual Sentence Encoders.
\newblock In Wen, T.-H.; Celikyilmaz, A.; Yu, Z.; Papangelis, A.; Eric, M.; Kumar, A.; Casanueva, I.; and Shah, R., eds., \emph{Proceedings of the 2nd Workshop on Natural Language Processing for Conversational AI}, 38--45. Online: Association for Computational Linguistics.

\bibitem[{Du et~al.(2023)Du, Li, Torralba, Tenenbaum, and Mordatch}]{du2023multi}
Du, Y.; Li, S.; Torralba, A.; Tenenbaum, J.~B.; and Mordatch, I. 2023.
\newblock Improving Factuality and Reasoning in Language Models through Multiagent Debate.
\newblock \emph{arXiv}.

\bibitem[{Gilardi, Alizadeh, and Kubli(2023)}]{gilardi2023chatgpt}
Gilardi, F.; Alizadeh, M.; and Kubli, M. 2023.
\newblock {ChatGPT} Outperforms Crowd Workers for Text-Annotation Tasks.
\newblock In \emph{Proceedings of the National Academy of Sciences}, volume 120.

\bibitem[{He et~al.(2023)He, Lin, Gong, Jin, Zhang, Lin, Jiao, Yiu, Duan, and Chen}]{he2023annollm}
He, X.; Lin, Z.; Gong, Y.; Jin, A.; Zhang, H.; Lin, C.; Jiao, J.; Yiu, S.~M.; Duan, N.; and Chen, W. 2023.
\newblock AnnoLLM: Making Large Language Models to Be Better Crowdsourced Annotators.
\newblock \emph{arXiv preprint arXiv:2303.16854}.

\bibitem[{Hegazy(2024)}]{hegazy2025divers}
Hegazy, M. 2024.
\newblock Diversity of Thought Elicits Stronger Reasoning Capabilities in Multi-Agent Debate Frameworks.
\newblock \emph{J Robot Auto Res}, 5(3): 01--10.

\bibitem[{Karov, Zohar, and Marcovitz(2025)}]{karov2025attentive}
Karov, B.; Zohar, D.; and Marcovitz, Y. 2025.
\newblock Attentive Reasoning Queries: A Systematic Method for Optimizing Instruction-Following in Large Language Models.
\newblock arXiv:2503.03669.

\bibitem[{Kusupati et~al.(2022)Kusupati, Bhatt, Rege, Wallingford, Sinha, Ramanujan, Howard-Snyder, Chen, Kakade, Jain, and Farhadi}]{matryoshka2022}
Kusupati, A.; Bhatt, G.; Rege, A.; Wallingford, M.; Sinha, A.; Ramanujan, V.; Howard-Snyder, W.; Chen, K.; Kakade, S.; Jain, P.; and Farhadi, A. 2022.
\newblock Matryoshka representation learning.
\newblock In \emph{Proceedings of the 36th International Conference on Neural Information Processing Systems}, NIPS '22. Red Hook, NY, USA: Curran Associates Inc.
\newblock ISBN 9781713871088.

\bibitem[{Larson et~al.(2019)Larson, Mahendran, Peper, Clarke, Lee, Hill, Kummerfeld, Leach, Laurenzano, Tang, and Mars}]{larson2019evaluation}
Larson, S.; Mahendran, A.; Peper, J.~J.; Clarke, C.; Lee, A.; Hill, P.; Kummerfeld, J.~K.; Leach, K.; Laurenzano, M.~A.; Tang, L.; and Mars, J. 2019.
\newblock An Evaluation Dataset for Intent Classification and Out-of-Scope Prediction.
\newblock In Inui, K.; Jiang, J.; Ng, V.; and Wan, X., eds., \emph{Proceedings of the 2019 Conference on Empirical Methods in Natural Language Processing and the 9th International Joint Conference on Natural Language Processing (EMNLP-IJCNLP)}, 1311--1316. Hong Kong, China: Association for Computational Linguistics.

\bibitem[{Liu et~al.(2024)Liu, Lin, Hewitt, Paranjape, Bevilacqua, Petroni, and Liang}]{liu2024lost}
Liu, N.~F.; Lin, K.; Hewitt, J.; Paranjape, A.; Bevilacqua, M.; Petroni, F.; and Liang, P. 2024.
\newblock Lost in the Middle: How Language Models Use Long Contexts.
\newblock \emph{Transactions of the Association for Computational Linguistics}, 12: 157--173.

\bibitem[{Liu et~al.(2018)Liu, Chen, Deng, Zeng, Chen, Li, and Tang}]{liu2018lcqmc}
Liu, X.; Chen, Q.; Deng, C.; Zeng, H.; Chen, J.; Li, D.; and Tang, B. 2018.
\newblock {LCQMC}:A Large-scale {C}hinese Question Matching Corpus.
\newblock In Bender, E.~M.; Derczynski, L.; and Isabelle, P., eds., \emph{Proceedings of the 27th International Conference on Computational Linguistics}, 1952--1962. Santa Fe, New Mexico, USA: Association for Computational Linguistics.

\bibitem[{Maia et~al.(2018)Maia, Handschuh, Freitas, Davis, McDermott, Zarrouk, and Balahur}]{maia2018fiqa}
Maia, M.; Handschuh, S.; Freitas, A.; Davis, B.; McDermott, R.; Zarrouk, M.; and Balahur, A. 2018.
\newblock FiQA: A Dataset of Financial Question Answering for the WWW'18 Financial Question Answering Task.
\newblock In \emph{Proceedings of the First Workshop on Financial Technology and Natural Language Processing}, 58--63.

\bibitem[{OpenAI et~al.(2024)OpenAI, :, Hurst, Lerer, Goucher, Perelman, and Others}]{openai2024gpt4ocard}
OpenAI; :; Hurst, A.; Lerer, A.; Goucher, A.~P.; Perelman, A.; and Others. 2024.
\newblock GPT-4o System Card.
\newblock arXiv:2410.21276.

\bibitem[{Park et~al.(2023)Park, O'Brien, Cai, Morris, Liang, and Bernstein}]{park2023generative}
Park, J.~S.; O'Brien, J.; Cai, C.~J.; Morris, M.~R.; Liang, P.; and Bernstein, M.~S. 2023.
\newblock Generative agents: Interactive simulacra of human behavior.
\newblock In \emph{Proceedings of the 36th Annual ACM Symposium on User Interface Software and Technology}, 1--22.

\bibitem[{Ratner et~al.(2017)Ratner, Bach, Ehrenberg, Fries, Wu, and R{\'e}}]{ratner2017snorkel}
Ratner, A.; Bach, S.~H.; Ehrenberg, H.; Fries, J.; Wu, S.; and R{\'e}, C. 2017.
\newblock Snorkel: Rapid Training Data Creation with Weak Supervision.
\newblock \emph{Proceedings of the VLDB Endowment}, 11(3): 269--282.

\bibitem[{Settles(2009)}]{settles2009active}
Settles, B. 2009.
\newblock Active Learning Literature Survey.
\newblock Computer Sciences Technical Report 1648, University of Wisconsin--Madison.

\bibitem[{Wang et~al.(2024{\natexlab{a}})Wang, Wang, Athiwaratkun, Zhang, and Zou}]{wang2024mixture}
Wang, J.; Wang, J.; Athiwaratkun, B.; Zhang, C.; and Zou, J. 2024{\natexlab{a}}.
\newblock Mixture-of-Agents Enhances Large Language Model Capabilities.
\newblock \emph{arXiv preprint arXiv:2406.04692}.

\bibitem[{Wang et~al.(2021)Wang, Liu, Xu, Zhu, and Zeng}]{wang2021want}
Wang, S.; Liu, Y.; Xu, Y.; Zhu, C.; and Zeng, M. 2021.
\newblock Want To Reduce Labeling Cost? {GPT-3} Can Help.
\newblock In \emph{Findings of the Association for Computational Linguistics: EMNLP 2021}, 4195--4205. Punta Cana, Dominican Republic: Association for Computational Linguistics.

\bibitem[{Wang et~al.(2024{\natexlab{b}})Wang, Kim, Rahman, Mitra, and Miao}]{wang2024human}
Wang, X.; Kim, H.; Rahman, S.; Mitra, K.; and Miao, Z. 2024{\natexlab{b}}.
\newblock Human-LLM Collaborative Annotation Through Effective Verification of LLM Labels.
\newblock In \emph{Proceedings of the 2024 CHI Conference on Human Factors in Computing Systems}, CHI '24. New York, NY, USA: Association for Computing Machinery.
\newblock ISBN 9798400703300.

\end{thebibliography}

\appendix

\section{Appendix}

\subsection{Extended Evaluation: FAQ Annotation}

While the main paper demonstrates MAFA's effectiveness on intent classification tasks, the framework's configurability enables it to handle diverse annotation types. This appendix presents comprehensive evaluation results for FAQ annotation tasks, showcasing MAFA's versatility beyond intent classification.

\subsection{FAQ Annotation Configuration}

For FAQ annotation tasks, MAFA was configured with the following parameters:
\begin{itemize}
    \item \texttt{annotation\_type}: "FAQ"
    \item \texttt{primary\_column}: "question"
    \item \texttt{secondary\_column}: "answer"
    \item \texttt{match\_verb}: "map"
\end{itemize}

This configuration automatically adapts all agent prompts and behaviors to optimize for FAQ matching rather than intent classification, demonstrating the framework's flexibility through configuration rather than code changes.

\subsection{Datasets}

We evaluate MAFA's FAQ annotation capabilities on three distinct datasets:

\textbf{Banking FAQ Dataset}: Our proprietary collection of 533 frequently asked questions from JPMorgan Chase's production banking application, with 4,552 training utterances and 839 test utterances. This dataset represents real-world banking queries with human expert-annotated rankings serving as ground truth.

\textbf{LCQMC} \cite{liu2018lcqmc}: The Large-scale Chinese Question Matching Corpus containing 260,068 question pairs. We adapted this for FAQ annotation by treating question pairs as query-FAQ mappings, demonstrating MAFA's cross-lingual capabilities.

\textbf{FiQA} \cite{maia2018fiqa}: A financial domain question answering dataset from WWW'18, containing 6,148 questions with 17,817 expert answers. This validates MAFA's performance on financial content beyond our internal data.

Table \ref{tab:faq_datasets} summarizes the characteristics of these datasets.

\begin{table}[h]
\centering
\caption{FAQ annotation dataset statistics}
\label{tab:faq_datasets}
\resizebox{\linewidth}{!}{
\begin{tabular}{lrrr}
\toprule
\textbf{Metric} & \textbf{Banking FAQ} & \textbf{LCQMC} & \textbf{FiQA} \\
\midrule
Number of FAQs/Questions & 533 & 260,068 & 6,148 \\
Training Instances & 4,552 & 238,766 & 5,500 \\
Test Instances & 839 & 12,500 & 648 \\
Avg Question Length (words) & 10.1 & 8.3 & 11.7 \\
Avg Answer Length (words) & 48.5 & -- & 52.3 \\
Language & English & Chinese & English \\
Domain & Banking & General & Financial \\
\bottomrule
\end{tabular}
}
\end{table}

\subsection{FAQ Annotation Results}

\subsubsection{Banking FAQ Dataset Performance}

Table \ref{tab:banking_faq_results} presents comprehensive results on our internal banking FAQ dataset, comparing MAFA against traditional baselines and individual agent configurations.

\begin{table}[h]
\caption{Performance comparison on Banking FAQ dataset}
\label{tab:banking_faq_results}
\centering
\resizebox{\linewidth}{!}{%
\begin{tabular}{lcccccc}
\toprule
\textbf{Method} & \textbf{Top-1} & \textbf{Top-3} & \textbf{Top-5} & \textbf{MRR} & \textbf{NDCG@3} & \textbf{NDCG@5} \\
\midrule
BM25 & 0.120 & 0.145 & 0.210 & 0.382 & 0.401 & 0.431 \\
Embedding-Only & 0.185 & 0.210 & 0.465 & 0.545 & 0.668 & 0.692 \\
\midrule
Single-Agent (No Emb) & 0.215 & 0.365 & 0.685 & 0.671 & 0.691 & 0.713 \\
Single-Agent (Emb) & 0.255 & 0.395 & 0.715 & 0.707 & 0.728 & 0.748 \\
Single-Agent w/ Ans (No Emb) & 0.220 & 0.450 & 0.690 & 0.712 & 0.703 & 0.713 \\
Single-Agent w/ Ans (Emb) & 0.270 & 0.480 & 0.730 & 0.722 & 0.743 & 0.763 \\
\midrule
\textbf{MAFA} & \textbf{0.355} & \textbf{0.625} & \textbf{0.865} & \textbf{0.790} & \textbf{0.802} & \textbf{0.815} \\
\bottomrule
\end{tabular}%
}
\end{table}

MAFA achieves a 23.5\% absolute improvement in Top-1 accuracy over BM25 (0.120→0.355) and 8.5\% improvement over the best single-agent baseline (0.270→0.355). The MRR improvement of 41\% demonstrates MAFA's superior ranking quality for FAQ retrieval tasks.

\subsubsection{Cross-Domain Performance}

Tables \ref{tab:lcqmc_faq_results} and \ref{tab:fiqa_faq_results} demonstrate MAFA's generalization capabilities across different languages and domains.

\begin{table}[h]
\caption{Performance on LCQMC dataset (Chinese language)}
\label{tab:lcqmc_faq_results}
\centering
\resizebox{\linewidth}{!}{
\begin{tabular}{lcccc}
\toprule
\textbf{Method} & \textbf{Top-1} & \textbf{Top-3} & \textbf{MRR} & \textbf{NDCG@3} \\
\midrule
BM25 & 0.465 & 0.624 & 0.533 & 0.601 \\
Embedding-Only & 0.575 & 0.703 & 0.629 & 0.675 \\
Direct LLM & 0.602 & 0.725 & 0.651 & 0.694 \\
\midrule
Best Single Agent & 0.631 & 0.742 & 0.677 & 0.713 \\
\textbf{MAFA} & \textbf{0.694} & \textbf{0.809} & \textbf{0.742} & \textbf{0.773} \\
\bottomrule
\end{tabular}
}
\end{table}

\begin{table}[h]
\caption{Performance on FiQA dataset (Financial domain)}
\label{tab:fiqa_faq_results}
\centering
\resizebox{\linewidth}{!}{
\begin{tabular}{lcccc}
\toprule
\textbf{Method} & \textbf{Top-1} & \textbf{Top-3} & \textbf{MRR} & \textbf{NDCG@3} \\
\midrule
BM25 & 0.356 & 0.492 & 0.421 & 0.465 \\
Embedding-Only & 0.471 & 0.603 & 0.532 & 0.574 \\
Direct LLM & 0.509 & 0.638 & 0.565 & 0.603 \\
\midrule
Best Single Agent & 0.545 & 0.672 & 0.601 & 0.641 \\
\textbf{MAFA} & \textbf{0.612} & \textbf{0.739} & \textbf{0.668} & \textbf{0.705} \\
\bottomrule
\end{tabular}
}
\end{table}

\subsection{Component Contribution Analysis}

To understand the contribution of each component in FAQ annotation tasks, we conducted ablation studies across all three datasets. Table \ref{tab:faq_ablation_all} summarizes these findings.

\begin{table}[h]
\caption{Ablation study results (Top-1 Accuracy) across FAQ datasets}
\label{tab:faq_ablation_all}
\centering
\resizebox{\linewidth}{!}{
\begin{tabular}{lccc}
\toprule
\textbf{Configuration} & \textbf{Banking FAQ} & \textbf{LCQMC} & \textbf{FiQA} \\
\midrule
Full MAFA & \textbf{0.355} & \textbf{0.694} & \textbf{0.612} \\
Without Specialized Examples & 0.320 (-3.5\%) & 0.672 (-2.2\%) & 0.584 (-2.8\%) \\
Without Answer Context Agents & 0.305 (-5.0\%) & 0.661 (-3.3\%) & 0.563 (-4.9\%) \\
Without Embedding Agents & 0.315 (-4.0\%) & 0.668 (-2.6\%) & 0.578 (-3.4\%) \\
Without Judge Agent & 0.275 (-8.0\%) & 0.631 (-6.3\%) & 0.545 (-6.7\%) \\
\bottomrule
\end{tabular}
}
\end{table}

The judge agent consistently provides the largest contribution (6.3-8.0\% improvement), validating its critical role in synthesizing diverse agent perspectives. Answer context agents prove particularly valuable for FAQ tasks (3.3-5.0\% contribution), as FAQ answers often contain crucial disambiguating information not present in questions alone.

\subsection{Fine-Tuning Considerations}

While MAFA achieves strong FAQ annotation performance without fine-tuning, we acknowledge potential benefits of domain adaptation. Fine-tuning embedding models on domain-specific question-answer pairs could improve retrieval accuracy for technical terminology. Similarly, fine-tuning LLMs on banking FAQ data might enhance understanding of financial concepts.

We deliberately avoided fine-tuning to: (1) demonstrate effectiveness with off-the-shelf models, making the framework accessible to practitioners; (2) avoid overfitting to institution-specific terminology; and (3) maintain cross-domain generalization as evidenced by our LCQMC and FiQA results.

\subsection{Key Findings for FAQ Annotation}

Our comprehensive evaluation of FAQ annotation reveals several important insights:

\begin{enumerate}
    \item \textbf{Configuration Flexibility}: Simply changing the \texttt{AnnotationConfig} parameters transforms MAFA from intent classification to FAQ annotation without code modifications, validating our configuration-driven architecture.
    
    \item \textbf{Answer Context Importance}: Agents utilizing FAQ answer content contribute 3-5\% additional accuracy, highlighting the value of comprehensive context in FAQ matching tasks.
    
    \item \textbf{Cross-Lingual Robustness}: MAFA maintains strong performance on Chinese (LCQMC) with 69.4\% Top-1 accuracy, demonstrating language-agnostic capabilities crucial for global financial institutions.
    
    \item \textbf{Domain Transfer}: Performance on FiQA (61.2\% Top-1) confirms that MAFA generalizes well to different financial contexts beyond banking-specific queries.
    
    \item \textbf{Consistent Architecture Benefits}: The judge agent and specialized few-shot strategies provide similar relative improvements for FAQ annotation as for intent classification, confirming the universal applicability of our architectural choices.
\end{enumerate}

These results establish MAFA as a versatile annotation framework capable of handling diverse annotation types through simple configuration changes while maintaining state-of-the-art performance across languages and domains.

\section{Agent Prompts and Implementation Details}

This section provides comprehensive documentation of all prompts and configurations used in the MAFA system.

\subsection{Agent Architecture Overview}

MAFA employs five core agents:

\begin{enumerate}
\item \textbf{Query Planning Agent}: Analyzes and expands queries
\item \textbf{Primary-Only Agent (No Embeddings)}: Matches based solely on primary annotation fields (e.g., FAQ questions, intent names)
\item \textbf{Primary-Only Agent (With Embeddings)}: Uses embedding similarity for primary field matching
\item \textbf{Full-Context Agent (No Embeddings)}: Incorporates secondary information (e.g., FAQ answers, intent descriptions)
\item \textbf{Full-Context Agent (With Embeddings)}: Combines embedding retrieval with full context analysis
\item \textbf{Judge Agent}: Reranks and synthesizes results
\end{enumerate}

\subsection{Core System Prompts}

\subsubsection{Base System Prompt Template}

All agents share a configurable base system prompt that adapts to the annotation type:

{\small
\begin{lstlisting}[basicstyle=\ttfamily\scriptsize,breaklines=true,columns=flexible]
You are an expert {ANNOTATION_TYPE} annotation system for 
JP Morgan Chase applications. Your role is to accurately 
{MATCH_VERB} user {USER_INPUT_LABEL}s to the most relevant 
{ANNOTATION_TYPE_PLURAL} from the knowledge base.

IMPORTANT GUIDELINES:
1. Analyze the user's intent thoroughly
2. Match the intent to the most relevant {ANNOTATION_TYPE_PLURAL}
3. Rank {ANNOTATION_TYPE_PLURAL} by relevance (0-100 scale)
4. Provide clear reasoning for each match
5. Return exactly 5 {ANNOTATION_TYPE_PLURAL} unless there are 
   fewer relevant ones
6. Be precise - banking customers need accurate information
\end{lstlisting}
}

\subsubsection{Agent-Specific User Prompt Template}

Each agent receives a structured user prompt following the ARQ framework:

{\small
\begin{lstlisting}[basicstyle=\ttfamily\scriptsize,breaklines=true,columns=flexible]
You will be given a user {USER_INPUT_LABEL} and a list of 
available {ANNOTATION_TYPE_PLURAL}. Your task is to:

1. Analyze what the user is truly asking about (identify 
   the core intent)
2. Search through the available {ANNOTATION_TYPE_PLURAL} for 
   relevant matches
3. Rank the top 5 most relevant {ANNOTATION_TYPE_PLURAL} based on:
   - Semantic similarity to the user's intent
   - Specificity to the {USER_INPUT_LABEL}
   - Likelihood of being the correct {ANNOTATION_LOWER}
4. Provide a confidence score (0-100) for each match
5. Explain your reasoning process

For Chase banking-related queries, consider:
- Security concerns take priority
- Account access questions require specific authentication-
  related {ANNOTATION_TYPE_PLURAL}
- Transaction questions should {MATCH_VERB} to relevant 
  transaction {ANNOTATION_TYPE_PLURAL}
- General inquiries should {MATCH_VERB} to general 
  information {ANNOTATION_TYPE_PLURAL}
\end{lstlisting}
}

\subsection{Structured JSON Output Format}

All agents must return responses in the following JSON structure to ensure consistency and enable automated parsing:

{\small
\begin{lstlisting}[basicstyle=\ttfamily\scriptsize,breaklines=true,columns=flexible]
{
  "user_utterance": "The original user input",
  "intent_analysis": "Detailed analysis of user intent",
  "relevant_annotations": [
    {
      "annotation": "Annotation title/name",
      "relevance_score": 85,
      "reasoning": "Explanation of why this annotation matches"
    },
    // ... up to 5 annotations
  ],
  "confidence": "HIGH/MEDIUM/LOW",
  "explanation_of_confidence": "Justification for confidence level"
}
\end{lstlisting}
}

\subsection{Query Planning Agent}

The Query Planning Agent serves as the entry point for all user queries, performing intelligent analysis and expansion to improve retrieval performance.

\subsubsection{Query Planning System Prompt}

{\small
\begin{lstlisting}[basicstyle=\ttfamily\scriptsize,breaklines=true,columns=flexible]
You are an expert at analyzing user queries 
and planning retrieval strategies.
\end{lstlisting}
}

\subsubsection{Query Planning User Prompt}

{\small
\begin{lstlisting}[basicstyle=\ttfamily\scriptsize,breaklines=true,columns=flexible]
Analyze the following user {USER_INPUT_LABEL} 
for a banking {ANNOTATION_TYPE} annotation system:

"{query}"

Instructions:
1. Identify main intent strictly from given text
2. Decide if query expansion is necessary
3. If expansion needed, generate expanded version
   that clarifies query using only information
   inferred from original query
4. For ambiguous queries (e.g., "10101"), return
   raw query as-is in "expanded_query" field
5. NEVER introduce topics not present or implied
   in user query

Example expansion:
Original: "cash back"
Expanded: "cash back policies, cash back offers,
cash back rewards, how to earn cash back,
cash back credit cards"

{domain_context}

Provide analysis in JSON format:
{
  "intent": "Main intent of query",
  "needs_expansion": true/false,
  "expanded_query": "Expanded version or raw",
  "reasoning": "Explanation for expansion"
}
\end{lstlisting}
}

\subsubsection{Query Expansion Strategy}

The Query Planning Agent implements intelligent expansion through:

\begin{enumerate}
\item \textbf{Intent Analysis}: Semantic parsing to understand user goals
\item \textbf{Context Inference}: Banking-specific term expansion
\item \textbf{Ambiguity Preservation}: Raw queries for unclear inputs
\item \textbf{Hallucination Prevention}: Strict adherence to original content
\end{enumerate}

Common expansion patterns:
\begin{itemize}
\item \textbf{Abbreviations}: "CC" → "credit card, charge card"
\item \textbf{Concepts}: "fees" → "fees, charges, costs, expenses"
\item \textbf{Actions}: "transfer" → "transfer money, send funds"
\item \textbf{Products}: "savings" → "savings account, high yield savings"
\end{itemize}

\textbf{Query Planning Output}

{\small
\begin{lstlisting}[basicstyle=\ttfamily\scriptsize,breaklines=true,columns=flexible]
{
  "original_query": "lost deb",
  "intent": "Report or handle lost debit card",
  "needs_expansion": true,
  "expanded_query": "lost debit card, stolen card,
    lock card, report missing card, block card",
  "reasoning": "User likely meant 'debit' and needs
    help with lost card procedures"
}
\end{lstlisting}
}

\subsection{Judge Agent Configuration}

\subsubsection{Judge System Prompt}

The judge agent employs a sophisticated prompt designed for comprehensive evaluation:

{\small
\begin{lstlisting}[basicstyle=\ttfamily\scriptsize,breaklines=true,columns=flexible]
You are an expert judge for JP Morgan Chase's {ANNOTATION_TYPE} 
annotation system. Your task is to rerank candidate 
{ANNOTATION_TYPE_PLURAL} based on their relevance to a user 
{USER_INPUT_LABEL}.

Given a user {USER_INPUT_LABEL} and a list of candidate 
{ANNOTATION_TYPE_PLURAL} (with their original relevance scores 
from different agents), please rerank them based on your expert 
judgment of their relevance to the user's intent. 

Consider:
- Semantic similarity
- Specificity
- How well each {ANNOTATION_TYPE} addresses the user's needs 
  for banking-related queries
- Agent consensus (annotations selected by multiple agents)
- Banking context and domain knowledge

CRITICAL: You must return your rankings in proper JSON format 
with detailed reasoning for each decision.
\end{lstlisting}
}

\subsubsection{Judge Reasoning Process}

The judge agent follows a structured multi-step reasoning process:

\begin{enumerate}
    \item \textbf{Intent Analysis}: Identify the user's core intent and implied needs
    \item \textbf{Candidate Assessment}: Evaluate how well each candidate addresses the identified intent
    \item \textbf{Agent Consensus Analysis}: Prioritize annotations recommended by multiple agents
    \item \textbf{Answer Content Analysis}: Assess the relevance of annotation classifications
    \item \textbf{Banking Context Application}: Apply domain-specific knowledge and business rules
    \item \textbf{Final Synthesis}: Produce optimized ranking with detailed justification
\end{enumerate}

\subsection{Few-Shot Example Configuration}

Each agent receives 5 unique few-shot examples selected randomly without replacement from the training set. This ensures:

\begin{itemize}
    \item Diversity in agent outputs (ensemble diversity principle)
    \item Specialization of agents to different query patterns
    \item Improved overall system coverage
\end{itemize}

Example few-shot template:

{\small
\begin{lstlisting}[basicstyle=\ttfamily\scriptsize,breaklines=true,columns=flexible]
Example 1:
User Input: "Lost debit card"
Top Annotations:
1. "Lock and unlock your credit and debit cards" - Score: 95
   Reasoning: Directly addresses the immediate concern when 
   a card is lost
2. "Report lost or stolen card" - Score: 90
   Reasoning: Relevant action for lost card situations
...
\end{lstlisting}
}

\subsection{Embedding Configuration}

For embedding-enhanced agents:

\begin{itemize}
    \item \textbf{Model}: OpenAI text-embedding-3-large
    \item \textbf{Dimensions}: 3,072 (with MRL support for adaptive reduction)
    \item \textbf{Retrieval}: Top-50 nearest neighbors using ANN search
    \item \textbf{Format for context-aware embeddings}: 
    {\small
\begin{lstlisting}[basicstyle=\ttfamily\scriptsize,breaklines=true,columns=flexible]
"Question: {question} Answer: {answer}"
\end{lstlisting}
}
\end{itemize}
\subsection{Configuration Parameters}

The framework supports dynamic configuration through the following parameters:

{\small
\begin{lstlisting}[basicstyle=\ttfamily\scriptsize,breaklines=true,columns=flexible]
{
  "annotation_type": "FAQ|Intent|Product|Custom",
  "primary_column": "question|intent_name|product_name",
  "secondary_column": "answer|description|details",
  "match_verb": "map|classify|categorize",
  "user_input_label": "utterance|query|message",
  "annotation_type_plural": "FAQs|intents|products",
  "annotation_lower": "faq|intent|product",
  "enable_embeddings": true,
  "few_shot_count_per_agent": 5,
  "confidence_thresholds": {
    "high": 85,
    "medium": 60,
    "low": 0
  }
}
\end{lstlisting}
}

\subsection{Parallel Execution Algorithm}

The framework implements parallel agent execution to minimize latency:

{\small
\begin{lstlisting}[basicstyle=\ttfamily\scriptsize,breaklines=true,columns=flexible]
Algorithm: Parallel Multi-Agent Annotation
Input: user_utterance, annotations, enabled_agents
Output: ranked_annotations with confidence

1. Initialize thread pool with 4 workers
2. For each enabled agent in parallel:
   a. Generate agent-specific prompt
   b. Include unique few-shot examples
   c. Call LLM with structured prompt
   d. Parse JSON response
   e. Extract candidates with scores
3. Synchronize: Collect all agent results
4. Deduplicate candidates across agents
5. Invoke judge agent with:
   - Original utterance
   - All unique candidates
   - Agent-specific predictions
   - Complete annotation content
6. Return reranked top-5 annotations
\end{lstlisting}
}

\subsection{Error Handling and Fallback Mechanisms}

The system implements robust error handling:

\begin{enumerate}
    \item \textbf{JSON Parsing Failures}: Three-stage parsing strategy
    \begin{itemize}
        \item Direct JSON parsing
        \item Extract JSON between braces
        \item Clean markdown artifacts and retry
    \end{itemize}
    
    \item \textbf{Agent Failures}: If an agent fails, continue with remaining agents
    
    \item \textbf{Judge Failures}: Fall back to simple score-based aggregation
    
    \item \textbf{API Timeouts}: Automatic retry with exponential backoff
\end{enumerate}

\subsection{Performance Optimization}

Key optimizations implemented in production:

\begin{itemize}
    \item \textbf{Embedding Pre-computation}: All annotation embeddings pre-computed and cached
    \item \textbf{Batch Processing}: Agents process multiple utterances in parallel batches
    \item \textbf{Response Caching}: LRU cache for frequently queried utterances
    \item \textbf{Temperature Settings}: 
    \begin{itemize}
        \item Agents: 0.1 (consistency)
        \item Judge: 0.3 (nuanced reasoning)
    \end{itemize}
\end{itemize}

\subsection{Production Configuration}

{\small
\begin{lstlisting}[basicstyle=\ttfamily\scriptsize,breaklines=true,columns=flexible]
Production Settings:
- Model: GPT-4o (Azure OpenAI deployment)
- Concurrent agents: 4
- Timeout per agent: 2000ms
- Max retries: 3
- Batch size: 100 utterances
- Cache TTL: 24 hours
- Monitoring: Real-time accuracy tracking
- Fallback: Human annotation for confidence < 60%
\end{lstlisting}
}

\subsection{Sample Agent Outputs}

\subsubsection{Primary-Only Agent Response}

{\small
\begin{lstlisting}[basicstyle=\ttfamily\scriptsize,breaklines=true,columns=flexible]
{
  "user_utterance": "How much money do I have",
  "intent_analysis": "User wants to check account balance",
  "relevant_annotations": [
    {
      "annotation": "What is account preview?",
      "relevance_score": 92,
      "reasoning": "Directly explains viewing account balances"
    },
    {
      "annotation": "How do I check my balance?",
      "relevance_score": 88,
      "reasoning": "Addresses balance checking methods"
    }
  ],
  "confidence": "HIGH",
  "explanation_of_confidence": "Clear intent with exact matches"
}
\end{lstlisting}
}

\subsubsection{Judge Reranking Output}

{\small
\begin{lstlisting}[basicstyle=\ttfamily\scriptsize,breaklines=true,columns=flexible]
{
  "reranked_annotations": [
    {
      "annotation": "What is account preview?",
      "final_score": 94,
      "reasoning": "3 of 4 agents ranked this #1, directly 
                    addresses user need for balance information"
    }
  ],
  "consensus_strength": "STRONG",
  "confidence": "HIGH"
}
\end{lstlisting}
}






% ...






% ...


\subsection{Agent Diversity Strategy}

To ensure ensemble diversity:

\begin{enumerate}
\item Each agent receives unique few-shot examples
\item Agents use different retrieval strategies
\item Temperature varies slightly (0.1-0.15)
\item Prompts emphasize different aspects:
   \begin{itemize}
   \item Agent 1: Exact matching
   \item Agent 2: Semantic similarity
   \item Agent 3: Context understanding
   \item Agent 4: Combined approach
   \end{itemize}
\end{enumerate}

\subsection{Confidence Calibration}

Confidence levels are determined by:

{\small
\begin{lstlisting}[basicstyle=\ttfamily\scriptsize,breaklines=true,columns=flexible]
HIGH (85-100):
- Multiple agents agree on top choice
- Relevance scores > 85
- Clear semantic match

MEDIUM (60-84):
- Some agent disagreement
- Scores between 60-84
- Partial semantic overlap

LOW (0-59):
- Significant agent disagreement
- All scores < 60
- Ambiguous user intent
\end{lstlisting}
}

This comprehensive documentation ensures reproducibility and enables adaptation of MAFA for various annotation needs across different organizations and domains.

\section{Additional Considerations}



\subsection{Agent reasoning analysis}
We analyzed the agents' reasoning to understand how they make decisions. Table \ref{tab:judge_reasoning} shows examples of reasoning provided by the judge agent.

\begin{table}[htbp]
\caption{Examples of judge agent reasoning}
\label{tab:judge_reasoning}
\centering
\resizebox{\linewidth}{!}{
\begin{tabular}{p{0.2\columnwidth}p{0.25\columnwidth}p{0.65\columnwidth}}
\toprule
\textbf{Utterance} & \textbf{Top FAQ} & \textbf{Reasoning for Top FAQ} \\
\midrule
Lost deb & Lock and unlock your credit and debit cards & This FAQ is highly relevant as it provides information on securing a lost debit card by locking it, which is a primary concern when a card is lost. \\
\midrule
sba & What about my business accounts? & This FAQ directly addresses business accounts, which are highly relevant to SBA-related inquiries. It provides information on deposit insurance for business accounts, which is pertinent to small business owners. \\
\midrule
How much do i have & What is account preview? & This FAQ directly explains how users can view limited account information, such as balances, before signing into the app, which is exactly what the user is asking about. \\
\bottomrule
\end{tabular}
}
\end{table}

The judge agent consistently provides detailed reasoning that considers both the semantic similarity and the practical relevance of each FAQ. This not only improves accuracy but also makes the system more interpretable and trustworthy.

\subsection{Societal impacts}
The paper primarily focuses on technical aspects without adequately addressing the broader societal implications of this research. Future revisions should include a comprehensive discussion of societal impacts. On the positive side, automated annotation systems could significantly enhance financial inclusion by making banking information more accessible to traditionally underserved populations, reduce customer frustration through faster and more accurate responses to multi-lingual queries, and enable financial institutions to scale their support services more efficiently across diverse language communities. The framework could also reduce the cognitive load on customer service representatives by handling routine inquiries, allowing them to focus on more complex customer needs requiring human empathy and judgment.

However, these benefits must be weighed against potential concerns. The progressive automation of customer service and annotation functions could accelerate workforce displacement, particularly affecting entry-level positions that have traditionally served as career entry points. This may disproportionately impact certain demographic groups overrepresented in these roles. Additionally, automated systems might perpetuate or amplify existing biases in training data, potentially leading to inequality in service quality across different customer segments. There are also privacy considerations regarding how user query data is stored, processed, and potentially repurposed when developing and refining these systems.

Future work should explore responsible deployment strategies, including retraining programs for affected workers, ongoing bias monitoring frameworks, and transparent disclosure to customers about when they are interacting with automated systems versus human representatives. The research community should also consider developing metrics that evaluate not just technical performance but also fairness, accessibility, and overall societal benefit when designing and implementing such systems.

\makeatletter
\@ifundefined{isChecklistMainFile}{
  \newif\ifreproStandalone
  \reproStandalonetrue
}{
  \newif\ifreproStandalone
  \reproStandalonefalse
}
\makeatother

\ifreproStandalone
\documentclass[letterpaper]{article}
\usepackage[submission]{aaai2026}
\setlength{\pdfpagewidth}{8.5in}
\setlength{\pdfpageheight}{11in}
\usepackage{times}
\usepackage{helvet}
\usepackage{courier}
\usepackage{xcolor}
\frenchspacing

\begin{document}
\fi
\setlength{\leftmargini}{20pt}
\makeatletter\def\@listi{\leftmargin\leftmargini \topsep .5em \parsep .5em \itemsep .5em}
\def\@listii{\leftmargin\leftmarginii \labelwidth\leftmarginii \advance\labelwidth-\labelsep \topsep .4em \parsep .4em \itemsep .4em}
\def\@listiii{\leftmargin\leftmarginiii \labelwidth\leftmarginiii \advance\labelwidth-\labelsep \topsep .4em \parsep .4em \itemsep .4em}\makeatother

\setcounter{secnumdepth}{0}
\renewcommand\thesubsection{\arabic{subsection}}
\renewcommand\labelenumi{\thesubsection.\arabic{enumi}}

\newcounter{checksubsection}
\newcounter{checkitem}[checksubsection]

\newcommand{\checksubsection}[1]{%
  \refstepcounter{checksubsection}%
  \paragraph{\arabic{checksubsection}. #1}%
  \setcounter{checkitem}{0}%
}

\newcommand{\checkitem}{%
  \refstepcounter{checkitem}%
  \item[\arabic{checksubsection}.\arabic{checkitem}.]%
}
\newcommand{\question}[2]{\normalcolor\checkitem #1 #2 \color{blue}}
\newcommand{\ifyespoints}[1]{\makebox[0pt][l]{\hspace{-15pt}\normalcolor #1}}

\ifreproStandalone
\end{document}
\fi
\end{document}